\documentclass{article}

\usepackage{microtype}
\usepackage{graphicx}
\usepackage{subcaption}
\usepackage{booktabs} 

\usepackage{hyperref}

\usepackage[preprint]{icml2026}

\usepackage{amsmath}
\usepackage{amssymb}
\usepackage{mathtools}
\usepackage{amsthm}
\usepackage{bm}
\newcommand{\rvx}{\bm{x}}
\newcommand{\E}{\mathbb{E}}

\usepackage[capitalize,noabbrev]{cleveref}

\theoremstyle{plain}

\theoremstyle{definition}

\theoremstyle{remark}

\usepackage[textsize=tiny]{todonotes}

\newcommand{\passone}{\textit{pass@$1$}}
\newcommand{\passk}[1]{\textit{pass@$#1$}}

\newcommand{\OURMETHOD}{Order-Token Search }
\newcommand{\OURMETHODtext}{Order-Token Search}

\newcommand{\update}[1]{{\color{black}#1}}

\icmltitlerunning{Improving Diffusion LM Decoding via Joint Search}

\begin{document}

\twocolumn[
  \icmltitle{Improving Diffusion Language Model Decoding\\
  through Joint Search in Generation Order and Token Space}



  \icmlsetsymbol{equal}{*}
  \icmlsetsymbol{senior}{$\dagger$}

\begin{icmlauthorlist}
  \icmlauthor{Yangyi Shen}{stanford,equal}
  \icmlauthor{Tianjian Feng}{zju,equal}
  \icmlauthor{Jiaqi Han}{stanford}
  \icmlauthor{Wen Wang}{zju}
  \icmlauthor{Tianlang Chen}{stanford}
  \icmlauthor{Chunhua Shen}{zju}
  \\
  \icmlauthor{Jure Leskovec}{stanford,senior}
  \icmlauthor{Stefano Ermon}{stanford,senior}
\end{icmlauthorlist}

  \icmlaffiliation{stanford}{Department of Computer Science, Stanford University, California, USA}
\icmlaffiliation{zju}{Department of Computer Science, Zhejiang University, Zhejiang, China}

  \icmlcorrespondingauthor{Yangyi Shen}{pyyshen@stanford.edu}

  \icmlkeywords{Machine Learning, ICML}

  \vskip 0.3in
]



\printAffiliationsAndNotice{${}^*$Equal contribution. ${}^\dagger$Equal senior supervision.}
\begin{abstract}
Diffusion Language Models (DLMs) offer order-agnostic generation that can explore many possible decoding trajectories. 
However, current decoding methods commit to a single trajectory, limiting exploration in trajectory space.
We introduce \textbf{\OURMETHODtext} to explore this space through jointly searching over generation order and token values. 
Its core is a likelihood estimator that scores denoising actions, enabling stable pruning and efficient exploration of diverse trajectories. 
Across mathematical reasoning and coding benchmarks, \OURMETHOD consistently outperforms baselines on GSM8K, MATH500, Countdown, and HumanEval (3.1\%, 3.8\%, 7.9\%, and 6.8\% absolute over backbone), matching or surpassing \textit{diffu}-GRPO post-trained d1-LLaDA.
Our work establishes joint search as a key component for advancing decoding in DLMs.

\end{abstract}

\section{Introduction}
\label{sec:introduction}

\begin{figure*}[t]
    \centering
    \includegraphics[width=1.0\linewidth]{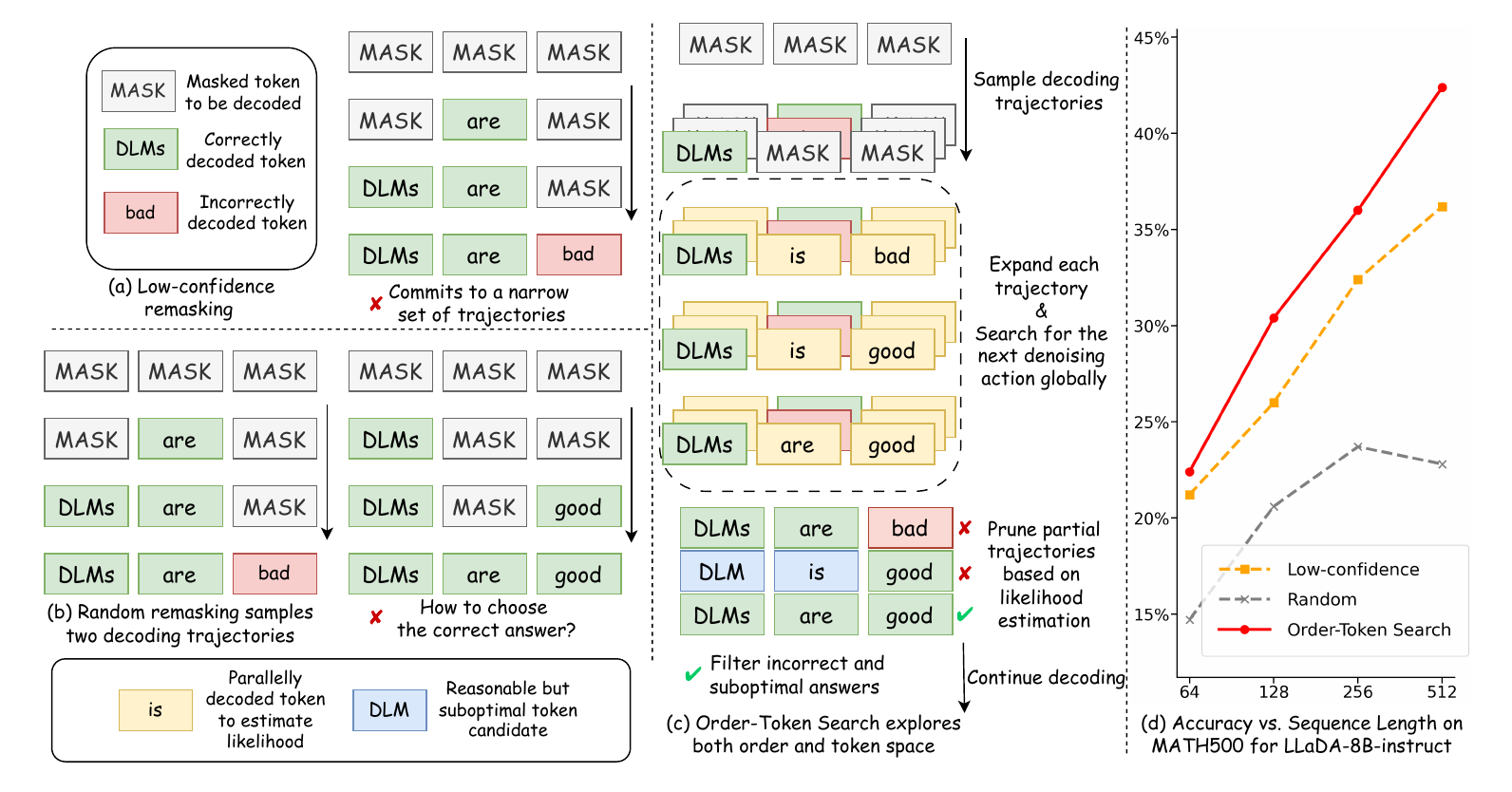}
    \caption{\textbf{Why heuristic remasking fails, and how \OURMETHOD helps.} 
(a) \textbf{Low-confidence remasking} greedily fixes high-confidence tokens early and only revises low-confidence positions, which effectively commits to a narrow set of decoding trajectories and can get stuck in an incorrect completion. 
(b) \textbf{Random remasking} uniformly samples different remasking patterns, producing multiple distinct trajectories (different orders and token assignments), but offers no principled way to select the correct trajectory among competing candidates. 
(c) \textbf{\OURMETHOD} maintains beams of partial trajectories, periodically expands each trajectory by proposing denoising actions that change both which position to update (order) and what token to write (token), and then prunes partial trajectories using a likelihood estimation function to filter incorrect and suboptimal candidates before continuing decoding. 
(d) \textbf{MATH500 accuracy plot} (example dataset) illustrating \OURMETHODtext’s advantage over low-confidence and random remasking.
}
    \label{fig:search_illustration}
\end{figure*}

Diffusion Language Models (DLMs) have emerged as a promising alternative to autoregressive models for sequence generation. In particular, Masked Diffusion Models (MDMs) \citep{sahoo2024simple,shi2024simplified} trains on a core objective: learning to reconstruct original text by denoising sequences corrupted with random masks at varying rates. At inference, generation starts from a fully masked sequence and proceeds over multiple denoising steps: the model predicts token distributions for all positions, samples tokens for masked positions, and re-masks a subset of tokens to construct the next input. 
One key opportunity is that MDMs enable order-agnostic decoding: they revise many positions in parallel, allowing adaptive generation orders rather than left-to-right. This flexibility makes the choice of generation order a central design decision for decoding.

However, current decoding methods produce only a single trajectory per run. In MDMs, common decoders follow heuristic remasking rules that effectively determine the trajectory, which positions are updated at each step and with what tokens. For instance, \textbf{random remasking} mirrors the training corruption by repeatedly masking a random subset of tokens, whereas \textbf{low-confidence} remasking greedily fixes high-confidence tokens as context and only revises uncertain ones \citep{llada,kim2025train}. As illustrated in Figure~\ref{fig:search_illustration}(a)(b), because these heuristics neither branch over alternative generation orders nor maintain competing intermediate states, a single run can easily miss the specific order–token trajectory needed for a correct solution.

Empirically, existing methods occupy opposite ends of an exploration--exploitation trade-off.
The \passk{k} metric captures the probability that at least one of $k$ samples is correct.
\textbf{Low-confidence remasking} \emph{exploits} by committing early to high-confidence tokens and restricting decoding to a narrow set of trajectories, which boosts \passone~but leads to slow \passk{k} growth as $k$ increases.
In contrast, \textbf{random remasking} \emph{explores} a much broader set of trajectories and attains higher \passk{k}, but achieves weak \passone~because it cannot reliably select good ones.
Thus, prior methods either over-exploit or explore blindly, leaving much of the model's multi-trajectory potential unrealized. We address this with explicit trajectory search and pruning.

Here we propose \textbf{\OURMETHODtext}, a decoding algorithm that performs search in the joint space of generation orders and token values. As shown in Figure~\ref{fig:search_illustration}(c), \OURMETHOD maintains beams of partially denoised states and periodically expands each beam by proposing multiple denoising actions, then prunes using a stable likelihood estimation function that reliably scores partial trajectories. Across mathematical reasoning and coding benchmarks, \OURMETHOD consistently improves \passone~over prior best decoding, matching or surpassing the gains of \textit{diffu-}GRPO post-training \citep{zhao2025d1} (e.g., +3.1/3.8/7.9/6.8\% absolute over low-confidence remasking on GSM8K/MATH500/Countdown/HumanEval). Figure~\ref{fig:search_illustration}(d) illustrates the gains on MATH500: across sequence lengths, \OURMETHOD consistently outperforms both low-confidence and random remasking.

Our contributions are threefold:
\textbf{(1)} We identify an exploration--exploitation trade-off in MDM decoding through \passk{k}: low-confidence remasking over-exploits, improving \passone{} but suppressing \passk{k}, while random remasking blindly explores, boosting \passk{k} but degrading \passone{} (Section~\ref{sec:scaling}).
\textbf{(2)} We propose \OURMETHODtext, a decoding algorithm that searches in the joint space of generation orders and token values and stably prunes partial trajectories with a likelihood estimator (Section~\ref{sec:method}).
\textbf{(3)} We demonstrate consistent gains across mathematical reasoning and coding benchmarks, matching or surpassing improvements typically obtained from post-training (Section~\ref{sec:experiments}).

\section{Background}
\label{sec:background}

This section establishes the technical foundation for our work. We review fundamentals of Masked Diffusion Models, formalize key concepts for remasking strategies, and define our evaluation metrics for the exploration trade-off.

\subsection{Discrete Diffusion Models}
\label{subsec:mdm}
Diffusion Language Models extend discrete diffusion modeling to text, and recent large-scale training has made them competitive with autoregressive models \citep{llada,dream2025}. 
Concretely, discrete diffusion models adapt the forward diffusion process and the reverse denoising process \citep{pmlr-v37-sohl-dickstein15,NEURIPS2020_4c5bcfec,NEURIPS2019_3001ef25, song2021scorebased} to discrete data by establishing the diffusion process over a discrete domain $\mathbf{x} \in \mathcal{X}$, where $\mathbf{x}$ is a one-hot vector denoting tokens from a vocabulary of size $|\mathcal{X}|$ \citep{austin2021structured}. Given a prior $\bm\pi$, the forward process $q$ incrementally corrupts the original data $\mathbf{x}_0$ into a target prior distribution $\text{Cat}(\cdot;\bm\pi)$. Over continous time $t \in [0,1]$, it forms a sequence of increasingly noisy latent variables $\mathbf{x}_t$, through the conditional marginal distribution $q(\mathbf{x}_t \mid \mathbf{x}_0) = \text{Cat}\left(\mathbf{x}_t; \alpha_t \mathbf{x}_0 + (1 - \alpha_t) \bm\pi\right)$. 
Here, $\alpha_t$ is a monotonically decreasing noise schedule that satisfies boundary conditions $\alpha_0 = 1$ and $\alpha_1 = 0$. Furthermore, we can achieve the transition probability between any two intermediate time points $0 < s < t < 1$ through $q(\mathbf{x}_t \mid \mathbf{x}_s,\mathbf{x}_0) = \text{Cat}\left(\mathbf{x}_t; \alpha_t/\alpha_s \mathbf{x}_s + (1 - \alpha_t/\alpha_s) \bm\pi\right)$.

Masked Diffusion Model (MDM), a specific instance of this framework, utilizes the prior $\bm\pi = \bm m$ to achieve absorbing-state diffusion, a particularly suitable setting for language modeling \citep{sahoo2024simple,shi2024simplified,lou2023discrete}. Here, $\bm m$ is a one-hot vector corresponding to a special MASK token. Defining $s$ as the time step immediately preceding $t$, the posterior distribution simplifies to:
{\small
\begin{equation}
q(\mathbf{x}_s \mid \mathbf{x}_t, \mathbf{x}_0) =
\begin{cases}
    \text{Cat}\left(\mathbf{x}_s; \mathbf{x}_t \right), & \mathbf{x}_t \neq \bm m \\
    \text{Cat}\left(\mathbf{x}_s; \frac{\alpha_s - \alpha_t}{1-\alpha_t}\mathbf{x}_0 + \frac{1-\alpha_s}{1-\alpha_t}\bm m \right), & \mathbf{x}_t \neq \bm m \\
\end{cases}
\label{eq:posterior}
\end{equation}
}
The reverse (denoising) process is modeled by $p_\theta(\mathbf{x}_s\mid\mathbf{x}_t)=q(\mathbf{x}_s \mid \mathbf{x}_t, \mathbf{x}_\theta (\mathbf{x}_t))$, where $p_\theta$ is a parameterized distribution that reverses $q$, and $\mathbf{x}_\theta (\mathbf{x}_t)$ denotes a neural network trained to predict the original clean data $\mathbf{x}_0$ from its noisy version $\mathbf{x}_t$. This network is optimized by minimizing the negative evidence lower bound, thereby learning to approximate the true posterior distribution.

\subsection{Remasking Strategies in MDM}
\label{subsec:remasking}

In masked generative models, decoding starts from a fully masked sequence, $\mathbf{x}_1 = (\text{MASK}, \ldots, \text{MASK})$. The model then iteratively refines this sequence over a series of steps. At each step, the model predicts logits for all currently masked positions. The critical action in this reverse process is the transfer of a prediction—that is, the act of replacing a selected \text{MASK} token with its predicted value, thereby committing to that prediction for subsequent steps. The rule that determines which masked token to transfer next is known as the \textbf{remasking strategy}, and it defines the decoding order. We focus on three primary strategies:

\textbf{Random Remasking.} The strategy used during training. The next position to unmask is chosen uniformly at random from the set of all remaining masked positions. This is a baseline that ensures unbiased, order-agnostic generation.
\textbf{Low-Confidence Remasking.} A common inference-time strategy. The position with the highest confidence score is unmasked next; the positions with lower scores are remasked. Formally, at each step, the model computes a confidence score for position $i$ as the predicted probability assigned to the currently selected token at that position $s_i = p_\theta(\cdot \mid \mathbf{x}_t)_i$. The position with the \textit{maximum} score $s_i$ is unmasked. The intuition is to resolve the token position where the model has the greatest certainty first, potentially mitigating error propagation \citep{llada,kim2025train}.
\textbf{Autoregressive (AR).} We force the MDM to keep the leftmost predicted token and remask all following tokens. This baseline decouples the effect of generation order and solely examines the effect of diverse token selection.

\subsection{Evaluation Metrics for Exploration Trade-off}
\label{subsec:bg_metrics}

Evaluating generative models on the trade-off between exploration and exploitation requires metrics that capture both deterministic performance and the model's inherent capability. We use the following standard metrics established in prior work \citep{yue2025rlreasoning}: 
\textbf{Accuracy (\textit{pass@$\mathbf{1}$})}, the probability that a single generated sample is correct. This is the primary metric for evaluating one-trial performance and represents the expected accuracy when using the model in a deterministic setting. 
\textbf{\textit{pass@$\mathbf{k}$}}, the probability that at least one sample out of $k$ independent generations is correct. This metric estimates the model's inherent capability to solve a problem given sufficient sampling budget. For a problem with $n$ generated samples of which $c$ are correct, it is estimated as: $\passk{k} \approx 1 - \binom{n - c}{k}/\binom{n}{k}$.

The gap between \passone\ and \passk{k} measures how much a decoding method benefits from sampling. 
If \passone\ is strong but \passk{k} increases slowly with $k$, the generated samples are highly correlated, indicating conservative decoding that commits early and explores few distinct trajectories. 
Conversely, if \passone\ is weak but \passk{k} rises quickly with $k$, the model can reach correct solutions under sampling, but the decoding method cannot reliably select them. 
Our goal is to translate this sampling-driven capability into higher \passone\ by explicitly exploring and pruning in the joint space of generation orders and token choices.

\begin{figure*}[t]
\centering
\includegraphics[width=1.0\linewidth]{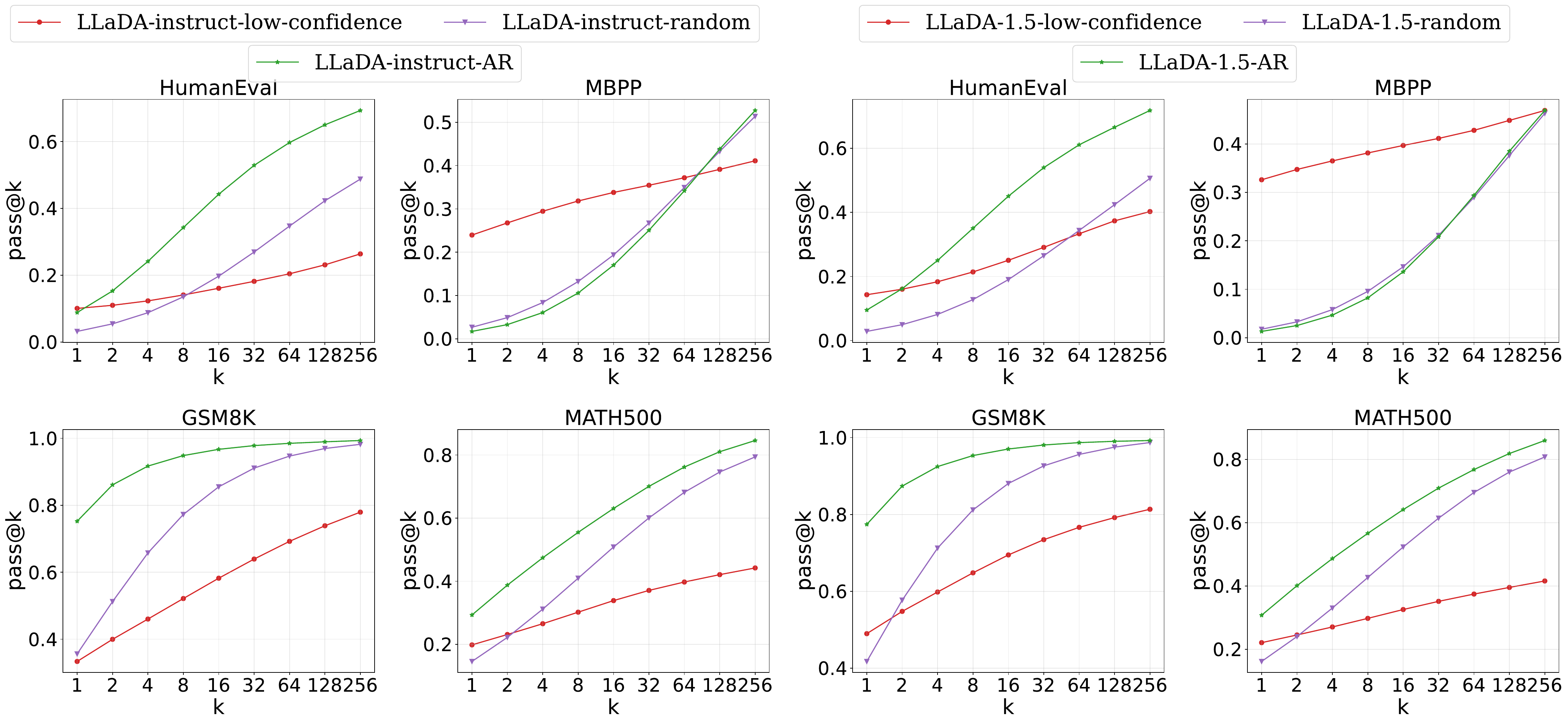}
\caption{\textbf{Standard DLM decoders reveal an exploration–exploitation trade-off.} 
Empirical \passk{k} curves for LLaDA-8B-Instruct and LLaDA-1.5 on mathematical reasoning and coding benchmarks. 
Low-confidence remasking often achieves higher \passone{} by exploiting local confidence, but its \passk{k} grows slowly with $k$, indicating limited coverage from a narrow set of trajectories.
In contrast, random remasking (order diversity) and AR decoding (token diversity) typically start with lower \passone{} yet reach much higher \passk{k} by $k=256$, reflecting broader solution-space coverage through more diverse trajectories.
We use \passk{k} here only as a diagnostic of latent multi-trajectory potential; the goal is to convert this potential into improved \passone{} via structured exploration and selection.}
\label{fig:passk_curves}
\end{figure*}

\section{Trade-off in Standard Decoding Methods}
\label{sec:scaling}
In this section, we study the exploration--exploitation trade-off among common DLM decoding methods. Specifically, we ask \textit{how the diversity of sampled decoding trajectories impacts problem-solving performance}. We compare three representative approaches that emphasize different sources of diversity: low-confidence remasking (greedy, low diversity), random remasking (high diversity in generation order), and autoregressive decoding (high diversity in token selections).
This comparison is particularly valuable because DLMs offer unique flexibility in generation order compared to autoregressive models, yet optimal strategies for leveraging this flexibility remain unclear.

\textbf{Low-confidence remasking: higher \passone{} but slower \passk{k} growth.} Figure~\ref{fig:passk_curves} summarizes our key findings. Using LLaDA models \citep{llada,llada1.5} trained for flexible-order generation, we evaluate each decoding method in terms of single-sample accuracy (\passone) and multi-sample coverage (\passk{k}) on mathematical reasoning (GSM8K, MATH500) and coding (HumanEval, MBPP) benchmarks, to understand how diversity in generation order and token selection affects performance.
Across 8 dataset--model settings, low-confidence remasking achieves the best \passone{} in 4 and the second-best in 3 (e.g., improving MBPP by 0.2--0.3 absolute).
This competitive \passone{} performance is consistent with the intuition that committing first on high-confidence tokens provides stable context and reduces error propagation within a single trajectory. However, as the sampling budget $k$ increases, \passk{k} for low-confidence remasking grows more slowly and is eventually overtaken by other baselines. This is likely because low-confidence remasking over-exploits local confidence and narrows the set of feasible decoding trajectories.

\textbf{Exploratory approaches: generally lower \passone{} but faster \passk{k} growth.} In contrast, random remasking and left-to-right autoregressive (AR) order typically start from a lower \passone{}, but benefit much more from additional samples: \passk{k} keeps rising with $k$ and eventually surpasses low-confidence remasking across most settings (e.g., on MATH500 at \passk{256}, random/AR reach $\sim$0.8 while low-confidence ends near $\sim$0.4). 
This behavior suggests that diversity enables the model to uncover correct solutions that are missed by a single committed trajectory. Such diversity can arise either from varying generation orders (random remasking) or from varying token selections under a fixed order (AR decoding with temperature).
We also observe that AR decoding performs best on mathematical reasoning, achieving the highest \passone{} and \passk{k} on GSM8K and MATH500. A plausible explanation is that these datasets more closely follow left-to-right, step-by-step solution structure, making a fixed causal order advantageous. In contrast, this pattern does not extend to coding: on HumanEval, AR attains the lowest \passone. Overall, these results underscore the importance of diversity in both generation order and token choices for achieving broad coverage across tasks.

\textbf{Summary and implication.} Standard decoding methods expose a clear exploration–exploitation trade-off in DLMs. Greedy methods such as low-confidence remasking exploit local confidence to optimize \passone{}, but do so by collapsing decoding to a narrow set of trajectories, leaving many valid solutions unexplored. More exploratory approaches, in contrast, reveal the model’s ability to solve additional problems by exploring alternative generation orders or token choices, though this comes at the cost of lower per-sample reliability. Importantly, our use of \passk{k} serves only to diagnose this latent potential for exploration—not as a target metric. The central challenge, therefore, is how to leverage the benefits of diversity to increase single-sample accuracy. 

\begin{figure*}[t]
    \centering
    \includegraphics[width=1.0\linewidth]{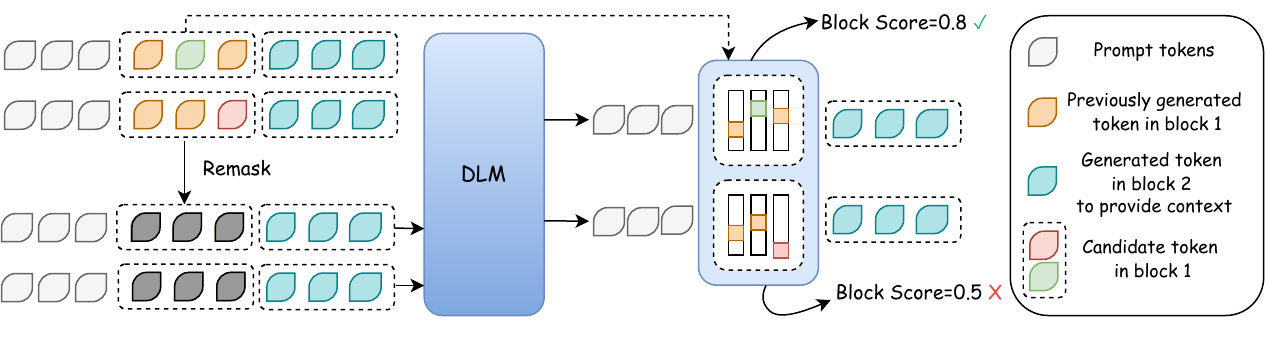}
    \caption{Illustration of a pruning stage in Order‑Token Search for DLMs. 
    At a search step, we have $2$ fully denoised sequences (on the leftmost), with yellow tokens unmasked in previous steps.
    \textbf{We then mask the current block (the middle $3$ tokens) and    measure its likelihood through feeding each masked candidate into the DLM to obtain each token's probability.}
    The score function computes the chain-rule product of token probabilities and prunes the lower-likelihood candidate.}
    \label{fig:prune_illustration}
\end{figure*}

\section{Method: Order-Token Search for Diffusion Language Models}
\label{sec:method}
Section~\ref{sec:scaling} shows that improving \passone{} requires converting trajectory diversity into a reliable single sample. To this end, we introduce \OURMETHODtext, a novel algorithm inspired by beam search but designed for the parallel, order-agnostic dynamics of DLMs. \OURMETHOD performs search in the joint space of generation order and token choices, and uses the model’s likelihood predictions to score partial trajectories and stably prune unpromising ones. This section presents its two key components—search and prune—with full pseudocode in Appendix~\ref{app:algo}.

\subsection{Search Process}

\OURMETHOD maintains a beam set of $K$ partially denoised candidates, initialized as $K$ identical copies of
$\mathbf{x}_1 = [\mathbf{c};\text{MASK}^L]$, where $\mathbf{c}$ is the prompt and $\text{MASK}^L$ denotes $L$ mask tokens.
Decoding proceeds over continuous denoising time $t\in[0,1]$. Unlike standard one-trajectory sampling, which fixes a single sequence of remasking decisions, our goal is to explicitly branch over \emph{both} sources of uncertainty: \textbf{what} tokens to write and \textbf{when/where} to write them (i.e., the induced generation order).

Concretely, at a user-specified interval $(s,t)$, we expand each beam independently for a number of denoising steps.
During expansion, the MDM produces token distributions for all currently masked positions, and the algorithm proposes multiple denoising actions that correspond to different joint samples from the order-token space (e.g., different subsets of positions to update and different sampled token assignments at those positions).
Each proposed action yields a new candidate state, thereby creating multiple partial trajectories from the same prefix.
This branching step is key for \OURMETHOD to explore diverse trajectories, while downstream pruning (Section~\ref{sec:pruning}) keeps only the top-$K$ candidates according to a stable likelihood estimate.

To manage computational complexity, we organize expansions using \textbf{block diffusion} \citep{block-diffusion,llada}.
Branching at every denoising step introduces an $O(K\cdot |T|)$ overhead over $|T|$ total steps. Instead, we only perform expansion at the boundaries of contiguous token blocks, and run independent denoising within each block.
This reduces the number of decision points from $|T|$ to $|B|$, where $|B|$ is the total number of blocks, yielding $O(K\cdot |B|)$ search overhead while still allowing substantial diversity in both generation order and token values.
After processing all blocks, we select a single output sequence from the remaining $K$ candidates based on the highest likelihood.
A detailed complexity analysis is provided in Appendix~\ref{app:complexity}.

\subsection{Pruning Process}
\label{sec:pruning}
The effectiveness of \OURMETHOD depends on a pruning criterion that can \emph{reliably} compare partially denoised candidates whose trajectories may differ in both token values and generation order. A direct choice is to score candidates using the standard MDM objective, i.e., the likelihood of reconstructing the full clean sequence from the current noisy state. However, this signal can be unstable in practice: when many tokens remain masked, the model is implicitly evaluated on extremely difficult infilling problems that are far from the ``small-step'' denoising behavior it learns well from training \citep{kim2025train}, and these scores can become high-variance and poorly calibrated for ranking.

\textbf{Scoring incremental denoising actions.}
Our key idea is to score a candidate \emph{by the specific denoising actions that produced it}, rather than by a monolithic full-sequence likelihood at an intermediate state. Consider a transition from a more corrupted state $\rvx_t$ to a less corrupted state $\rvx_s$ with $0 \le s < t \le 1$. Let $\rvx_0 \sim p_\theta(\rvx_0 \mid \rvx_t)$ denote a sample from the model’s conditional distribution over clean sequences given $\rvx_t$. We define an action-level score
\begin{equation}
s(\rvx_{t};\rvx_s) 
= \E_{\rvx_0 \sim p_\theta(\rvx_0\mid\rvx_{t})} 
\log p_\theta\left(\rvx_0 \,\middle|\, b(\rvx_s, \rvx_t, \rvx_0)\right),
\label{eq:likelihood-estimate}
\end{equation}
where $p_\theta(\cdot|\cdot)$ is the parametrized posterior from Section~\ref{subsec:mdm}. The function $b$ extracts the newly revealed blocks between $t$ and $s$, i.e.,
\update{$\{ i \mid \rvx_{t,i}=\text{MASK} \ \cap\ \rvx_{s,i}\neq \text{MASK}\}$,}
then masks exactly these positions in $\rvx_0$ and returns the resulting masked sequence. Intuitively, $s(\rvx_t;\rvx_s)$ measures how confidently the model can re-predict \emph{only the tokens that were newly denoised in this step}, using the surrounding context supplied by the model's own full prediction $\rvx_0$. Figure~\ref{fig:prune_illustration} visualizes this block-level scoring for one transition.

\begin{table*}[t]
\caption{\textbf{Model Performance on Mathematics and Coding Benchmarks.} We report accuracy across four benchmarks and multiple generation lengths for two base models (LLaDA and LLaDA-1.5). Bolded values indicate the best performance, and underlined values indicate the second best. \OURMETHOD achieves the highest average accuracy for both base models and attains the best dataset-level averages on MATH500, Countdown, and HumanEval, while remaining competitive with the strongest baselines on GSM8K.}
\label{tab:performance}
\resizebox{\linewidth}{!}{
\centering
\begin{tabular}{ll|*{21}{c}}
\toprule
 & & \multicolumn{5}{c}{GSM8K} & \multicolumn{5}{c}{MATH500} & \multicolumn{5}{c}{Countdown} & \multicolumn{5}{c}{HumanEval} & Average \\
\cmidrule(lr){3-7} \cmidrule(lr){8-12} \cmidrule(lr){13-17} \cmidrule(lr){18-22} \cmidrule(lr){23-23}
& Method / Seq Len & 64 & 128 & 256 & 512 & Avg & 64 & 128 & 256 & 512 & Avg & 64 & 128 & 256 & 512 & Avg & 64 & 128 & 256 & 512 & Avg &  \\
\midrule
LLaDA & Low-confidence    & 44.3 & 68.7 & 76.7 & 78.2 & 67.0 & 21.2 & 26.0 & 32.4 & 36.2 & 29.0 & 25.8 & 20.7 & 19.5 & 16.0 & 20.5 & 10.4 & 18.9 & 26.2 & 26.8 & 20.6 & 34.3 \\
      & Low-conf + MV     & 46.4 & 72.5 & 80.9 & 83.1 & \textbf{70.7} & 20.2 & 27.4 & 35.0 & 36.2 & 29.7 & 22.7 & 23.8 & 18.4 & 18.0 & 20.7 & 11.6 & 18.9 & 26.2 & 26.8 & 20.9 & 35.5 \\
      & Random remasking & 33.4 & 56.5 & 63.6 & 63.1 & 54.2 & 14.7 & 20.6 & 23.7 & 22.8 & 20.5 & 8.0 & 13.3 & 10.9 & 11.7 & 11.0 & 12.3 & 19.3 & 27.0 & 25.9 & 21.1 & 26.7 \\
      & Random + MV       & 43.1 & 70.7 & 80.2 & 80.3 & 68.6 & 17.2 & 26.2 & 31.8 & 31.8 & 26.8 & 6.3 & 15.2 & 14.1 & 15.2 & 12.7 & 15.9 & 17.1 & 29.9 & 28.1 & 22.7 & 32.7 \\
      & AR                & 34.0 & 62.7 & 75.7 & 76.9 & 62.3 & 18.8 & 23.4 & 27.4 & 34.4 & 26.0 & 10.6 & 12.9 & 13.3 & 14.1 & 12.7 & 10.4 & 20.1 & 29.9 & 31.7 & 23.0 & 31.0 \\
      & AR + MV           & 41.0 & 69.1 & 81.7 & 86.4    & 69.6    & 17.4 & 23.0 & 32.2 & 39.9 & 28.1 & 10.2 & 13.3 & 11.3 & 13.7 & 12.1 & 9.8  & 20.1 & 29.9 & 31.7 & 22.9 & 33.2    \\
      & AR + beam-search  & 40.3 & 70.4 & 81.1 & 82.9    &  68.7   & 22.2 & 26.6 & 35.4 & 39.8 & \underline{31.0} & 18.4 & 23.1 & 21.5 & 21.9    & \underline{21.2}    & 15.9 & 22.6 & 36.0 & 34.8 & \underline{27.3} & \underline{37.1}    \\
      & OTS All Blocks & 42.8 & 68.4 & 74.5 & 74.8 & 65.1 & 21.0 & 27.2 & 32.8 & 33.0 & 28.5 & 11.7 & 16.8 & 19.9 & 16.4 & 16.2 & 11.6 & 18.3 & 23.8 & 28.1 & 20.4 & 32.6 \\
      & OTS Future Blocks  & 43.3 & 66.8 & 76.2 & 78.9    &  66.3   & 22.1 & 28.4 & 31.6 & 36.8 & 29.7 & 15.2 & 18.0 & 20.3 & 16.0    & 17.4    & 12.8 & 16.5 & 28.1 & 31.7 & 22.3 & 33.9    \\
      & \textbf{Order-Token Search}& 45.6 & 71.6 & 79.8 & 83.3 & \underline{70.1} & 22.4 & 30.4 & 36.0 & 42.4 & \textbf{32.8} & 27.7 & 34.4 & 26.2 & 25.4 & \textbf{28.4} & 15.2 & 23.2 & 34.2 & 37.2 & \textbf{27.4} & \textbf{39.7} \\
\midrule
\midrule
LLaDA-1.5 & Low-confidence & 44.2 & 69.6 & 77.5 & 79.4 & 67.7 & 20.2 & 26.4 & 32.5 & 36.2 & 28.8 & 19.8 & 19.7 & 17.9 & 21.8 & 19.8 & 11.0 & 15.9 & 31.1 & 28.7 & 21.7 & 34.5 \\
          & \update{Low-conf + MV}  & 49.1 & 74.4 & 84.1 & 84.2 & \textbf{73.0} & 21.2 & 30.0 & 34.8 & 39.3 & \underline{31.3} & 20.7 & 23.8 & 20.3 & 25.4 & \underline{22.5} & 11.6 & 16.5 & 31.1 & 28.7 & 22.0 & \underline{37.2} \\
          & Random remasking & 45.0 & 60.6 & 65.6 & 66.9 & 59.5 & 19.6 & 21.1 & 27.9 & 28.8 & 24.4 & 8.5 & 11.8 & 10.1 & 11.4 & 10.5 & 11.8 & 17.6 & 26.8 & 26.5 & 20.7 & 28.8 \\
          & Random + MV    & 47.7 & 74.8 & 81.2 & 82.8 & 71.6 & 22.4 & 26.2 & 31.0 & 30.4 & 27.5 & 5.5 & 16.4 & 9.0 & 14.5 & 11.4 & 13.4 & 15.9 & 29.9 & 27.4 & 21.7 & 33.1 \\
          & \update{AR}             & 37.5 & 67.7 & 77.3 & 79.1 & 65.4 & 17.2 & 23.5 & 31.2 & 35.0 & 26.7 & 12.3 & 15.2 & 14.5 & 15.7 & 14.4 & 13.4 & 22.7 & 30.5 & 30.5 & 24.2 & 32.7 \\
          & \update{AR + MV}        & 40.7 & 72.9 & 83.8 & 84.2 & 70.4 & 18.4 & 26.4 & 33.6 & 37.9 & 29.1 & 12.5 & 17.2 & 14.8 & 16.0 & 15.1 & 12.8 & 22.6 & 30.5 & 30.5 & 24.1 & 34.7 \\
          & \update{AR + beam-search} & 45.1 & 73.2 & 82.0 & 84.9 & 71.3 & 19.0 & 26.4 & 35.2 & 38.8 & 29.9 & 14.8 & 21.1 & 16.0 & 20.3 & 18.1 & 12.2 & 20.7 & 32.8 & 33.8 & \underline{24.9} & 36.1 \\
          & OTS All Blocks & 45.6 & 70.7 & 78.9 & 78.3 & 68.4 & 20.8 & 26.4 & 31.8 & 34.6 & 28.4 & 9.0 & 21.5 & 20.7 & 22.3 & 18.4 & 13.4 & 16.6 & 25.6 & 26.8 & 20.6 & 34.0\\
          & OTS Future Blocks & 45.3 & 71.3 & 80.5 & 80.7 & 69.5 & 19.8 & 27.2 & 34.8 & 35.6 & 29.4 & 13.7 & 21.9 & 21.5 & 19.5 & 19.2 & 12.2 & 17.1 & 27.4 & 34.8 & 22.9 & 35.3 \\
          & \textbf{Order-Token Search} & 48.4 & 74.5 & 81.7 & 84.0 & \underline{72.2} & 24.4 & 30.8 & 37.4 & 42.4 & \textbf{33.8} & 27.7 & 31.3 & 23.8 & 29.3 & \textbf{28.0} & 15.2 & 22.6 & 36.0 & 36.6 & \textbf{27.6} & \textbf{40.4} \\
\bottomrule
\end{tabular}
}
\end{table*}

\textbf{Trajectory-level score for pruning.}
This action-level view yields a lower-variance and more comparable signal across candidates. Because each term focuses on a small set of tokens, it avoids penalizing an intermediate state for being ``globally hard'' when many masks remain, and instead evaluates whether a trajectory's past decisions (which order and what tokens it chose) are supported by the model. We assign each candidate a cumulative score by summing over the set of intervals $\mathcal{I}$ at which the algorithm performed branching and search:
$
S(\text{candidate}) = \sum_{(s,t)\in\mathcal{I}} s(\rvx_t;\rvx_s).
$
This sum captures the full generation history of partial trajectories: candidates that repeatedly make model-consistent denoising moves accumulate higher scores, while trajectories that rely on unlikely commitments are pruned early.

Overall, \OURMETHOD alternates between searching and pruning to turn trajectory diversity into improved accuracy (\passone{}). The search process maintains partial trajectories sampled from a joint distribution over which positions to reveal (generation order) and what tokens to write (token values). The pruning process then ranks these candidates using a trajectory-level score that decomposes likelihood into incremental denoising moves while providing a comprehensive measure of global coherence. By repeatedly keeping only the top-$K$ candidates after each search expansion, \OURMETHOD allocates a fixed inference-time compute budget toward exploring diverse order-token trajectories and reliably selecting the most coherent one.

\section{Experiments}
\label{sec:experiments}

We conduct a series of experiments to evaluate the effectiveness of \OURMETHOD (OTS) in improving the reasoning performance of MDMs. Our investigation centers on the following research questions:
(1) Does \OURMETHOD yield consistent improvements in reasoning accuracy over competitive baselines, such as low-confidence remasking and majority-voting, across a variety of tasks?
(2) How critical is \OURMETHODtext's dedicated likelihood estimator for pruning, compared to alternative scoring ablations?
(3) How does performance scale with NFE?
%

\subsection{Experimental Setup}
\label{subsec:exp-setup}

\textbf{Baselines.} We compare \OURMETHOD (OTS) against several strong baselines: 
\textbf{Low-confidence remasking}, a greedy decoding method adopted as an optimal base model configuration in \citet{zhao2025d1}.
\textbf{Random remasking}, an exploratory decoding method that re-masks uniformly at random.
\textbf{Random remasking with majority voting}, which generates a compute-equivalent set of diverse samples via random remasking and selects answers using a consistency heuristic \citep{majority-voting}.
\textbf{Low-confidence with majority voting}, which combines the greedy decoding with the consistency heuristic mentioned above.
\textbf{AR}, which follows the left-to-right autoregressive order in generation.
\textbf{AR with majority voting} and \textbf{AR with beam search}, which strengthen the AR baseline with, respectively, a consistency heuristic and a search based on OTS likelihood estimator.
\textbf{OTS All Blocks}, which scores each expansion by the joint log-likelihood of all blocks revealed so far;
\textbf{OTS Future Blocks}, which scores the newly-revealed blocks while masking all unrevealed future blocks (see Appendix~\ref{app:likelihood_ablation}).
\textbf{Order Search}, a computationally expensive algorithm that uses AR-style likelihood to search through the top-$K$ likely positions at every step; 
\textbf{Token Search}, an equally expensive algorithm that uses AR-style likelihood to search through the top-$K$ likely tokens for each position (see Appendix~\ref{app:ar_likelihood}). 

\textbf{Model and Tasks.} Our primary testbed is \textbf{LLaDA-8B-Instruct} \citep{llada}, a state-of-the-art open-source diffusion language model. Since it has not undergone post-training with methods like \textit{diffu}-GRPO \citep{zhao2025d1}, it offers a clean baseline for isolating the performance improvements attributable to our inference-time algorithm. \update{We additionally evaluate \textbf{LLaDA-1.5} \citep{llada1.5}, an RL post-trained variant of LLaDA, to verify that our conclusions hold even after reinforcement-learning-based post-training.}
For tasks, we evaluate on three mathematical reasoning and one coding benchmarks. \textbf{GSM8K} \citep{cobbe2021training} contains $\sim$1.32k grade school math problems requiring multi-step reasoning. \textbf{MATH500} \citep{math500} is a challenging subset of 500 high-school competition-level problems from the MATH \citep{hendrycksmath2021} dataset. \textbf{Countdown} \citep{tinyzero} is a combinatorial arithmetic game where the goal is to reach a target number using basic operations on a given set. \textbf{HumanEval} \citep{chen2021evaluating} is a code generation benchmark of 164 hand-written Python programming tasks, where models must synthesize a function from a docstring and pass unit tests.



\subsection{\OURMETHOD Improves Reasoning Accuracy}
\textbf{Overall performance: \OURMETHOD is the strongest decoding across benchmarks.}
As shown in Table~\ref{tab:performance}, for both LLaDA and LLaDA-1.5, \OURMETHOD achieves the highest average accuracy (39.7\% vs.\ 37.1\% for the best baseline on LLaDA, and 40.4\% vs.\ 37.2\% on LLaDA-1.5) and the best dataset-level averages on MATH500, Countdown, and HumanEval.
On GSM8K, alternative multi-sample strategies (Majority-Voting or AR with beam-search) trade wins with \OURMETHOD at specific sequence lengths, while \OURMETHOD remains consistently competitive.
Notably, \OURMETHOD also surpasses expensive post-training such as \textit{diffu-}GRPO \citep{zhao2025d1}; for example, at Seq Len 512 it improves GSM8K (83.3\% vs.\ 82.1\%), MATH500 (42.4\% vs.\ 40.2\%), and HumanEval (37.2\% vs.\ 34.8\%).

\textbf{Diffusion baselines: \OURMETHOD improves over remasking and voting on harder tasks.}
Within diffusion-style decoding, greedy low-confidence remasking is already strong, and adding majority voting consistently improves GSM8K averages for both models.
Random+MV, which replaces confidence-based remasking with random remasking, can be competitive on GSM8K but substantially degrades tasks requiring more structured reasoning: its Countdown averages drop to 12.7\% and 11.4\% for LLaDA and LLaDA-1.5, compared with 20.7\% and 22.5\% for Low-conf+MV.
In contrast, \OURMETHOD consistently outperforms these diffusion baselines on harder benchmarks: MATH500, Countdown, and HumanEval.
For example, OTS raises MATH500 averages to 32.8\% and 33.8\% (vs.\ 29.7\% and 31.3\% for Low-conf+MV), improves Countdown to 28.4\% and 28.0\%, and improves HumanEval to 27.4\% and 27.6\%.

\textbf{Autoregressive baselines: \OURMETHOD consistently outperforms AR.}
AR, AR+MV, and AR+beam-search are strong baselines on MATH500, Countdown, and HumanEval, with AR+beam-search achieving the second/third best average performance for both backbones.
Nevertheless, \OURMETHOD delivers clear gains.
Averaging dataset-level accuracies over both LLaDA variants, \OURMETHOD attains 71.2/33.3/28.2/27.5\% on GSM8K/MATH500/Countdown/HumanEval respectively, compared to 70.0/28.6/13.6/23.5\% for AR+MV and 70.0/30.5/19.7/26.1\% for AR+beam-search.
Notably, AR+beam-search is a $\texttt{block\_size}=1$ special case of our framework, yet it still underperforms full \OURMETHODtext, underscoring the value of jointly searching over generation orders and token choices rather than performing left-to-right beam search only in the token space.

\vspace{-5pt}
\subsection{The Necessity of Dedicated Likelihood Estimation}

\textbf{Robust incremental scoring is essential for pruning.}
Table~\ref{tab:performance} shows that even with the \emph{same} OTS branching and pruning procedure, the choice of likelihood estimator has a large impact.
Both OTS All Blocks and OTS Future Blocks underperform \OURMETHODtext.
For LLaDA, \OURMETHOD achieves 39.7\% overall accuracy, whereas OTS All Blocks and OTS Future Blocks drop to 32.6\% and 33.9\%, with the largest degradation on Countdown (28.4\% $\rightarrow$ 16.2\% / 17.4\%).
The same trend holds for LLaDA-1.5: \OURMETHOD reaches 40.4\% overall, versus 34.0\% / 35.3\% for the ablations, again with a large Countdown gap (28.0\% $\rightarrow$ 18.4\% / 19.2\%).
Together, these results validate our estimator in Eq.~\ref{eq:likelihood-estimate}: scoring incremental denoising actions while conditioning on the model's full-sequence predictive distribution yields a more stable and discriminative pruning signal than either joint history scoring within one forward pass or scoring without predicted future context.

\begin{table}[t]
\centering
\caption{\textbf{Accuracy of search algorithms with OTS or AR-style likelihood estimate (Seq Len 256).} Bolded values indicate best performance. OTS consistently outperforms both Order Search and Token Search that adopt an AR-style likelihood estimate.}
\label{tab:ar-likelihood}
\resizebox{0.98\linewidth}{!}{%
\begin{tabular}{lcccc}
\toprule
Decoding Method (Compute) & GSM8K & MATH500 & Countdown & HumanEval\\ 
\midrule
Token Search (3x) & 8.5 & 3.8 & 0.0 & 0.6\\
Order Search (3x) & 79.2 & 35.8 & 15.2 & 32.9\\
Order-Token Search (1x) & \textbf{79.8} & \textbf{36.0} & \textbf{26.2} & \textbf{34.2}\\
\bottomrule
\end{tabular}%
}
\end{table}

\textbf{AR-style likelihood is misaligned with MDMs.}
Search-based decoding depends on accurate likelihood estimates for pruning, but naive AR-style likelihood is mismatched to MDMs, which are trained to denoise \emph{multiple positions in parallel} (Section~\ref{subsec:mdm}).
In Table~\ref{tab:ar-likelihood}, Order Search and Token Search instantiate this AR-style estimate by scoring partially denoised sequences via forward passes conditioned only on the currently revealed tokens.
Both lag far behind \OURMETHODtext: on Countdown, \OURMETHOD outperforms Order Search by 11\%, while Token Search collapses to 0\% accuracy.
This indicates that directly porting AR beam-search heuristics to MDMs yields unreliable pruning signals; in contrast, \OURMETHOD succeeds by pairing joint order--token exploration with a likelihood estimation function tailored to incremental denoising.

\begin{figure}[t]
\centering
\includegraphics[width=\linewidth]{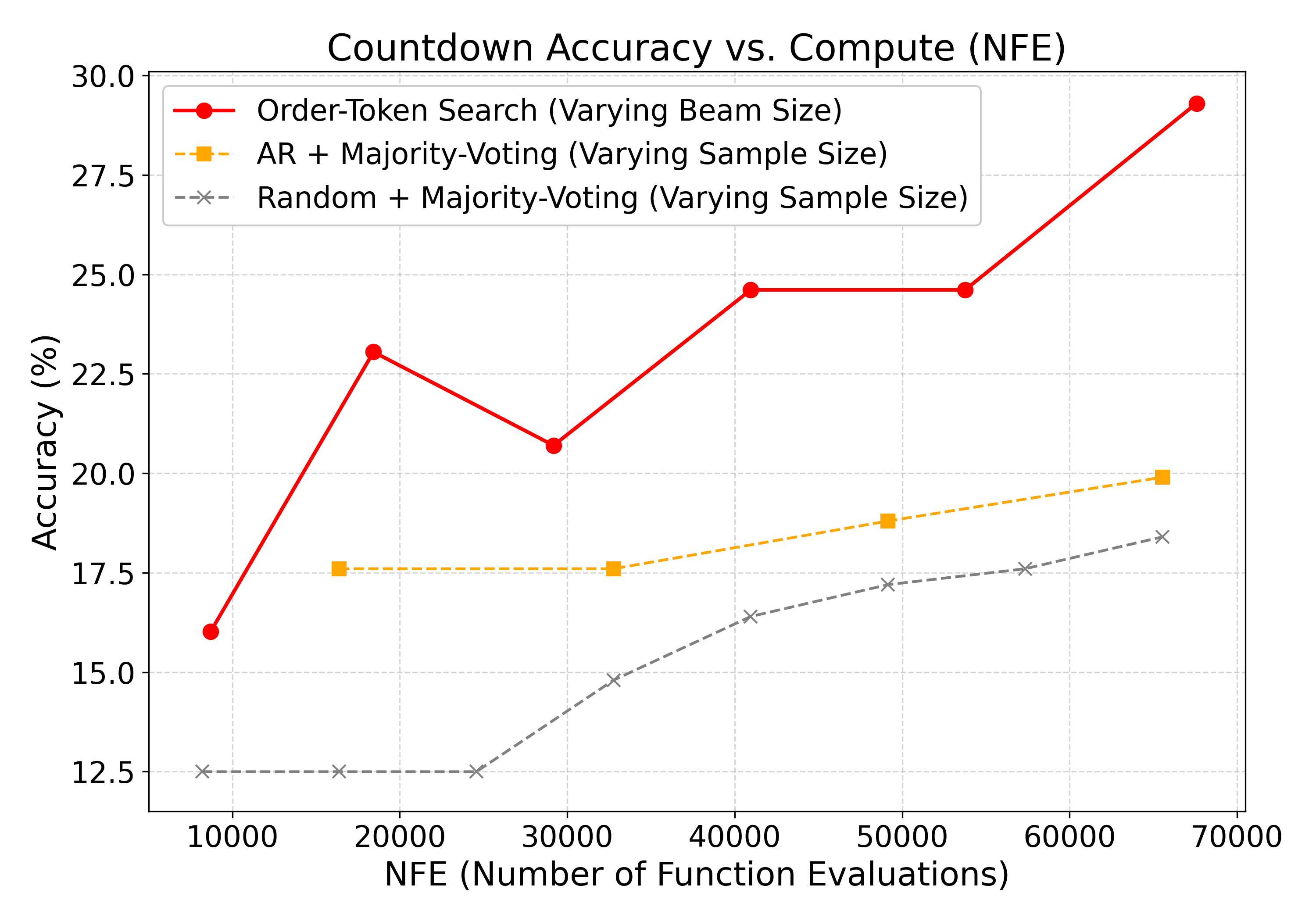}
\caption{\textbf{Countdown accuracy versus test-time compute (NFE)} for OTS and majority-voting baselines. For each method, we sweep beam size (OTS) or the number of samples (AR+MV, Random+MV), and set the largest configuration so the right-most points have approximately matched NFE. At this matched-compute point, OTS (beam size 6) achieves 29.3\% accuracy, versus 19.9\% for AR+MV and 18.4\% for Random+MV, demonstrating more efficient returns from additional test-time compute than drawing more independent samples.}
\label{fig:nfe-scale}
\end{figure}
\vspace{-5pt}

\subsection{\OURMETHOD Scaling with NFE}

\textbf{OTS scales more effectively with test-time compute than majority voting.}
We study scaling on Countdown by varying beam size for OTS and the number of samples for AR+MV and Random+MV under roughly matched FLOP budgets (Figure~\ref{fig:nfe-scale}).
At the matched-compute frontier, OTS with beam size 6 reaches 29.3\% accuracy, while AR+MV and Random+MV peak at 19.9\% and 18.4\%, respectively.
Moreover, OTS steadily improves with additional beams (16.0\% at beam 1 $\rightarrow$ 29.3\% at beam 6), whereas majority-voting methods show diminishing returns with more samples.
Overall, OTS more effectively converts extra NFE into accuracy gains by jointly searching over generation orders and token choices, rather than relying on heuristic sample aggregation over multiple independent trajectories.

\vspace{-5pt}
\section{Related Work}
\textbf{Diffusion Language Models.} Initial developments in discrete diffusion models were established by D3PM~\citep{austin2021structured} and further progressed using masked token approaches~\citep{sahoo2024simple, nie2024scaling}. Efficient versions such as Plaid~\citep{gulrajani2023likelihood} and SEDD~\citep{lou2023discrete} achieve performance comparable to GPT-2~\citep{radford2019language}, but their scalability still falls short of autoregressive models. The most recent scaling efforts include Dream~\citep{dream2025}, which adapts pre-trained autoregressive models into diffusion models, and LLaDA~\citep{llada}, which trains powerful diffusion language models from scratch.

\textbf{Test-Time Strategies.}
A primary method to enhance diffusion models is to increase test-time compute, often by using more denoising steps. Recent work has shown that expanding the inference-time sample space can guide generation toward high-reward outputs \citep{singhal2025generalframeworkinferencetimescaling,kim2025inference}, with techniques like re-masking being introduced to scale the denoising process for masked diffusion models specifically \citep{wang2025remasking}. In the broader context of language models, search algorithms like beam search, speculative decoding \citep{leviathan2023fast, xia2023speculative}, and contrastive decoding \citep{RGCL} have been developed to improve decoding beyond greedy selection. However, these algorithms are tailored for the autoregressive paradigm, where the search space is confined to token values given a fixed generation order.

Our work addresses this limitation. The parallel, order-agnostic decoding of MDMs creates a joint search space over both token values and their generation order, which is inaccessible to autoregressive methods. Our algorithm, \OURMETHODtext, is designed for this new paradigm, leveraging parallel decoding to explore multiple generation trajectories and select outputs based on overall likelihood.
\vspace{-5pt}
\section{Conclusion}
In this work, we revisited decoding for Diffusion Language Models (DLMs) through the lens of order--token trajectories: the intertwined decisions of where to denoise next and what to write. We identify a clear exploration--exploitation trade-off in standard decoding methods: low-confidence remasking over-exploits local confidence and commits early to a narrow set of greedy trajectories, while random remasking encourages broader trajectory diversity but lacks a reliable selection mechanism to choose good trajectories.

To reconcile this trade-off, we introduced \OURMETHODtext, which searches in the joint space of generation orders and token values. \OURMETHOD maintains partially denoised candidates, branches by proposing diverse denoising actions, and prunes with a likelihood estimator that scores incremental denoising decisions for comparison across competing trajectories. Across mathematical reasoning and coding benchmarks, this joint search yields systematic gains, matching or surpassing improvements typically obtained via post-training methods such as \textit{diffu}-GRPO. Overall, our results highlight joint order--token search as a key ingredient for converting DLMs' multi-trajectory capacity into reliable single-sample decoding performance.

\newpage

\section*{Impact Statement}
This paper presents work whose goal is to advance the field
of Machine Learning. There are many potential societal
consequences of our work, none which we feel must be
specifically highlighted here.

\bibliography{ref}
\bibliographystyle{icml2026}

\newpage
\appendix
\onecolumn
\clearpage
\appendix
\renewcommand\thesection{\Alph{section}}
\renewcommand\thefigure{S\arabic{figure}}
\renewcommand\thetable{S\arabic{table}}
\renewcommand\theequation{S\arabic{equation}}

\renewcommand\theHsection{appendix.\Alph{section}}
\renewcommand\theHfigure{S.\arabic{figure}}
\renewcommand\theHtable{S.\arabic{table}}
\renewcommand\theHequation{S.\arabic{equation}}

\setcounter{figure}{0}
\setcounter{table}{0}
\setcounter{equation}{0}

\section*{Use of LLMs}
Large language models (LLMs) were used solely to assist with grammar refinement and writing clarity during the manuscript preparation stage. All technical ideas, experimental designs, model implementations, and analyses were conceived and executed by the authors without reliance on LLMs. The use of LLMs did not influence research outcomes, data interpretation, or reported results. We carefully reviewed and edited all text to ensure accuracy, originality, and compliance with ethical and academic standards.

\section{Appendix}
\label{sec:appendix}

\subsection{\OURMETHOD Algorithm}
\label{app:algo}
To make our proposed decoding strategy more concrete, we present the pseudocode of our \hyperref[alg:search]{Order-Token Search}, which illustrates how partially masked sequences are expanded, scored and pruned, exploring both the token space and order space.

\begin{algorithm}[h]
\caption{Order-Token Search for Diffusion Language Models}
\label{alg:search}
\begin{algorithmic}[1]
\STATE \textbf{Input}: Prompt $\mathbf{p}$, model $p_\theta$, beam size $K$, generation length $L$, total steps $S$, search interval $N$, temperature $\tau$, number of blocks $b$.
\STATE Initialize beam set $\mathcal{B} \gets \{ (\mathbf{x}_i, \mathbf{s}[b], \text{score}) \}$, where $\mathbf{x}_i = [\mathbf{p}; \text{MASK}^{L}]$, $\mathbf{s}[0:b-1] = 0$, $\text{score}=0$ \COMMENT{$K$ identical beams}
\FOR{step $s \gets 1$ to $S$}
    \STATE $\mathbf{l} \gets p_\theta(\mathcal{B}.\mathbf{x})$ \COMMENT{Get logits for all beams, shape: $(K, L, V)$}
    \IF{$s \mod N == 0$} 
        \STATE $\mathcal{B}_{\text{candidates}} \gets \emptyset$
        \FOR{$(\mathbf{x}, \mathbf{s}, \text{score}) \in \mathcal{B}$}
            \STATE $\text{block\_idx} \gets \text{get\_current\_block\_index}(\textbf{x})$ \COMMENT{Compatible with semi-AR generation}
            \FOR{$i \gets 1$ to $K$} 
                \STATE $\tilde{\mathbf{l}} \gets \text{add\_gumbel\_noise}(\mathbf{l}_{\mathbf{x}}, \tau)$ 
                \STATE $\mathbf{x}_0 \gets \text{argmax}(\tilde{\mathbf{l}}, \text{dim}=-1)$ \COMMENT{Sample a candidate completion}
                \STATE $\mathbf{x}_{\text{candidate}} \gets \text{transfer\_tokens}(\mathbf{x}, \mathbf{x}_0, \mathbf{l}_{\mathbf{x}})$ \COMMENT{Only apply $\frac{L}{S}$ predicted tokens}
                \STATE $\mathbf{x}_{\text{full\_seq}} \gets \text{transfer\_all\_tokens}(\mathbf{x}, \mathbf{x}_0)$ \COMMENT{Apply all predicted tokens}
                \STATE $\mathbf{x}_{\text{masked}} \gets \text{mask\_tokens}(\mathbf{x}_{\text{full\_seq}}, \text{block\_idx})$ \COMMENT{Mask the current block}
                \STATE block\_score $\gets$ score\_block($\mathbf{x}_{\text{masked}}$, block\_idx) \COMMENT{Score the sequence}
                \STATE $\mathbf{s}[\text{block\_idx}] = \text{block\_score}$ 
                \STATE $\text{score} = \text{sum}( \mathbf{s}[0\!:\!\text{block\_idx}])$

                \STATE $\mathcal{B}_{\text{candidates}} \gets \mathcal{B}_{\text{candidates}} \cup \{ (\mathbf{x}_{\text{candidate}}, \mathbf{s}, \text{score}) \}$
            \ENDFOR
        \ENDFOR
        \STATE $\mathcal{B} \gets \text{top}_K(\mathcal{B}_{\text{candidates}})$ \COMMENT{Prune to the $K$ best candidates}
    \ELSE 
        \FOR{$(\mathbf{x}, \text{\_\_,\_\_}) \in \mathcal{B}$} 
            \STATE $\tilde{\mathbf{l}} \gets \text{add\_gumbel\_noise}(\mathbf{l}_{\mathbf{x}}, \tau)$
            \STATE $\mathbf{x}_0 \gets \text{argmax}(\tilde{\mathbf{l}}, \text{dim}=-1)$
            \STATE $\mathbf{x} \gets \text{transfer\_tokens}(\mathbf{x}, \mathbf{x}_0,\mathbf{l}_{\mathbf{x}})$ 
        \ENDFOR
    \ENDIF
\ENDFOR
\STATE \textbf{Return}: The sequence from $\mathcal{B}$ with the highest final score.
\end{algorithmic}
\end{algorithm}

\subsection{Alternative Likelihood Estimators for OTS}
\label{app:likelihood_ablation}

This appendix details two scoring ablations of \OURMETHOD (OTS) that isolate design choices in our robust likelihood estimator (Eq.~\ref{eq:likelihood-estimate}). Both baselines keep the \emph{same} OTS search/expansion procedure, and differ \emph{only} in the pruning score used to rank candidates.

\paragraph{Notation.}
Let $\rvx_t$ and $\rvx_s$ ($0\le s<t\le 1$) be the candidate states before/after an expansion, and let $\rvx_0 \sim p_\theta(\rvx_0 \mid \rvx_t)$ be the model's full-sequence prediction used as context.
We define three position sets:
\begin{align}
\Delta(s,t) &\coloneqq \{ i \mid \rvx_{t,i}=\text{MASK} \ \wedge\  \rvx_{s,i}\ne \text{MASK} \} \quad \text{(newly-revealed positions)}, \\
R(s) &\coloneqq \{ i \mid \rvx_{s,i}\ne \text{MASK} \} \setminus \text{prompt positions} \quad \text{(revealed-so-far)},\\
F(s) &\coloneqq \{ i \mid \rvx_{s,i}=\text{MASK} \} \quad \text{(unrevealed future positions)}.
\end{align}
Let $\text{mask}(\rvx_0, S)$ denote replacing positions in $S$ with $\text{MASK}$.

\subsubsection{OTS All Blocks: Joint Likelihood of All Revealed Blocks}
\label{app:all_blocks}

\textbf{Idea.} Instead of scoring only the newly-revealed blocks $\Delta(s,t)$ (Eq.~\ref{eq:likelihood-estimate}), this baseline scores the \emph{entire revealed prefix} $R(s)$ at each expansion, i.e., a harder joint infilling task.

\textbf{Score.} We replace Eq.~\ref{eq:likelihood-estimate} with:
\begin{equation}
s_{\textsc{all}}(\rvx_t;\rvx_s)
=
\E_{\rvx_0 \sim p_\theta(\rvx_0\mid \rvx_t)}
\log p\!\left(\rvx_0^{R(s)} \,\middle|\, \text{mask}(\rvx_0, R(s))\right),
\label{eq:score_all_blocks}
\end{equation}
Since $s_{\textsc{all}}(\rvx_t;\rvx_s)$ already evaluates \emph{all revealed blocks so far}, we use it directly as the pruning score at each search boundary (i.e., we do not sum over intervals).

\subsubsection{OTS Future Blocks: Likelihood Without Conditioning on Predicted Future}
\label{app:future_blocks}

\textbf{Idea.} Our main scorer conditions the likelihood of $\Delta(s,t)$ on the \emph{full} predicted $\rvx_0$, which includes predicted future content. This ablation removes that advantage by masking all unrevealed future positions $F(s)$ when scoring the current update.

\textbf{Score.} We score only the newly-revealed positions, but compute their likelihood using a context where future blocks are masked:
\begin{equation}
s_{\textsc{future}}(\rvx_t;\rvx_s)
=
\E_{\rvx_0 \sim p_\theta(\rvx_0\mid \rvx_t)}
\log p\!\left(\rvx_0^{\Delta(s,t)} \,\middle|\, \text{mask}(\rvx_0, \Delta(s,t)\cup F(s))\right).
\label{eq:score_future_blocks}
\end{equation}
This evaluates how well the model supports the current block using only the prompt and previously revealed blocks, without conditioning on any predicted future tokens.
The total candidate score is then accumulated over search-guided denoising steps,
$\sum_{(s,t)\in \mathcal{I}} s_{\textsc{future}}(\rvx_t;\rvx_s)$, matching the aggregation rule in \OURMETHOD.

\subsubsection{AR-style Likelihood: Left-to-Right Token Log-Likelihood}
\label{app:ar_likelihood}

For comparison, we describe the traditional likelihood used in autoregressive (AR) language models.
Given a partially generated sequence $\mathbf{x} = (x_1,\dots,x_L)$, an AR model factorizes the likelihood in a left-to-right manner:
\begin{equation}
p_{\textsc{ar}}(\mathbf{x}) = \prod_{i=1}^{L} p_\theta(x_i \mid x_{<i}),
\label{eq:ar_factorization}
\end{equation}
where $x_{<i} \coloneqq (x_1,\dots,x_{i-1})$.
Accordingly, the standard log-likelihood score for a candidate sequence is
\begin{equation}
s_{\textsc{ar}}(\mathbf{x}) = \sum_{i=1}^{L} \log p_\theta(x_i \mid x_{<i}).
\label{eq:ar_score}
\end{equation}

In AR decoding (e.g., beam search), candidates are ranked by accumulating token-level log-probabilities along the generation path. In our Order Search and Token Search baselines, we adopt an analogous AR-style score by summing log-likelihoods of the already revealed tokens in a forward pass.

\begin{table*}[t]
\centering
\caption{\update{Wall-clock time (in seconds) comparison on the Countdown dataset, averaged over all problems. This demonstrates that OTS (4 beams) is roughly 2-3x slower than a single heuristic-based remasking run, but about 2x faster than majority voting with 5 samples.}}
\label{tab:wall_time}
\begin{tabular}{lrrrr}
\toprule
Method / Generation length & 64 & 128 & 256 & 512 \\
\midrule
Heuristic-based remasking & 1.55 & 3.19 & 6.60 & 14.52 \\
+ Majority-voting (5 samples) & 7.73 & 15.94 & 32.99 & 72.59 \\
Order-Token Search (4 beams) & 3.52 & 7.46 & 16.64 & 40.41 \\
\bottomrule
\end{tabular}
\end{table*}

\subsection{Computation Complexity}
\label{app:complexity}

To manage computational complexity, we deliberately structure the search using \textbf{block diffusion} \citep{block-diffusion,llada}. This avoids the prohibitive cost of a naive search at every step, which would incur a complexity of $O(S \cdot K^2 \cdot L)$, where $S$ is the number of diffusion steps. The overhead of our approach is far lower. Let $L$ be the generation length, $K$ the beam size, and $B$ the number of blocks. The total Number of Function Evaluations (NFE) for OTS is the sum of $K$ independent denoising trajectories (costing $S \cdot K \cdot L$) and the likelihood evaluations for search, which are performed $B$ times at block boundaries (costing $B \cdot K^2 \cdot L$).

Thus, the total NFE is $\text{NFE(OTS)} \approx S \cdot K \cdot L + B \cdot K^2 \cdot L$. In our main experimental setting (where $S = L/2$ and $B = L/32$), with a typical beam size $K \approx 4$, this simplifies to $\text{NFE(OTS)} \approx (L^2 \cdot K)/2 + (K^2 \cdot L^2)/32 \approx 2.5 \cdot L^2$. This is critically important, as it is directly comparable to the NFE of a standard majority-voting baseline with 5 samples: $\text{NFE(MV-5)} = S \cdot 5 \cdot L = (L/2) \cdot 5 \cdot L = 2.5 \cdot L^2$. Therefore, OTS provides a structured joint search over the (order $\times$ token) space at roughly the same computational cost as a widely-used unstructured sampling baseline. After processing all blocks, the single best sequence is selected from the final $K$ candidates based on the highest likelihood. 
To further validate our analysis, we measured wall-clock time on the Countdown dataset, averaging over all problems and comparing heuristic-based remasking (e.g., low-confidence remasking), naive majority voting (5 samples), and OTS with 4 beams. As shown in Table \ref{tab:wall_time}, our optimized implementation of OTS runs faster than the majority-voting baseline.

\subsection{Comparison Study of Greedy Decoding and \OURMETHOD under Identical Temperature}
To further validate the effectiveness of our approach, we compare greedy decoding with Order-Token Search under identical temperature settings. This controlled setup rules out confounding factors and highlights the contribution of the search strategy itself. As reported in Table~\ref{tab:app-countdown-temp}, Order-Token Search delivers a remarkable performance boost on the Countdown dataset, confirming that the method provides tangible gains beyond simple greedy decoding.
\begin{table}[h]
\centering
\caption{Countdown task performance under different configurations.}
\begin{tabular}{c c}
\hline
\textbf{Seq Len($L$), Diffusion Steps($S$), Beam Size($K$), Temperature($T$)} & \textbf{Accuracy (\%)} \\
\hline
L=128,S=64,K=1,T=0.0 & 20.7 \\
L=128,S=64,K=1,T=0.4 & 22.7 \\
L=128,S=64,K=5,T=0.4 & 34.4 \\
\hline
\end{tabular}
\label{tab:app-countdown-temp}
\end{table}

\subsection{Additional Experimental Details}
\label{app:settings}
We provide further details on the experimental settings that complement the main results.


\subsubsection{Beam Search Settings}
Our \textbf{Order-Token Search} results (as shown in Table \hyperref[tab:performance]{1} and Table \hyperref[tab:ar-likelihood]{2} is configured with beam sizes of $K \in \{3,5,8\}$ and block size of $32$ along with a small search for the Gumbel noise temperature $\tau \in [0.2, 1.0]$, keeping in mind its role in balancing diversity and stability. As a general principle, a higher temperature introduces more diversity among the beams, but it can also risk destabilizing the token selection and decoding order. The settings used for our main experiments were chosen to maintain a reasonable balance between these factors.

In the main paper, we adopt the low-confidence remasking strategy together with the setting $\texttt{gen\_len} = 2 \times \texttt{diffusion\_steps}; \texttt{block\_size}=32$ for our baseline experiments. This configuration follows prior work \citep{zhao2025d1} and provides what can be regarded as a form of optimality: while it does not guarantee strict global optimality, it has been shown to yield a reasonably effective and competitive baseline under low-confidence conditions. Random-remasking majority-voting and \OURMETHOD both use the same configuration. \update{And we change \texttt{block\_size = gen\_len = 1} to simulate AR decoding on AR, AR majority-voting and AR Beam Search.}

For the Order Search and Token Search experiments reported in Table~\ref{tab:ar-likelihood}, 
we use the configuration with $K=3$. 

\subsubsection{Pass@k Evaluation Settings}
For pass@$k$ evaluation, we adopt the same configuration as in \citet{yue2025rlreasoning}. We set the temperature to $0.8$, which provides a balance between token diversity and plausibility. We use $\texttt{gen\_len} = \texttt{block\_size} = 256$, since the models we adopt are trained to generate sequences in a fully flexible order and we employ the same setup at inference time. For autoregressive decoding, we implement it via block diffusion with $\texttt{block\_size}=1$.

\subsubsection{Computational Cost Analysis}
Order Search generally search only on the decoding order of sequences based on the confidence of each position. This algorithm is designed based on the intuition that decoding order might change the ultimate accuracy. Therefore, at every step, we keep $K$ positions that have the highest probability from model logits independently unmasked. We then perform a look-ahead at the next step to have $K^2$ candidate sequences each with one more position unmasked. We calculate the confidence score and keep the top-$K$ candidates. In the experiment, we adopt the configuration of $K=3, T=0.0, \texttt{gen\_len}=256$ and the results also witness promising improvement. 

However, the algorithm is computationally expensive, for it requires $K^2 \times \texttt{gen\_len}$ forward passes in total. 
With $K=3$ and $\texttt{gen\_len}=256$, this amounts to $3^2 \times 256 = 2304$ forward evaluations. 
In contrast, our Order-Token Search with $K=5$ requires only 
$(128 \times 5) + \bigl(\tfrac{128}{32}\bigr) \times 25 = 740$ forward passes, 
where $128/32$ corresponds to the number of blocks and each block update involves $5 \times 5$ expansions. 

Token Search is closely related to Order Search, but instead of expanding $K$ positions at each step, 
it expands the top-$K$ most confident tokens for a single position. 
Starting from $K$ sequences, this again produces $K^2$ candidate sequences per step, 
leading to the same overall complexity of $K^2 \times \texttt{gen\_len}$ forward passes. 
For instance, with $K=3$ and $\texttt{gen\_len}=256$, Token Search also requires 
$3^2 \times 256 = 2304$ forward evaluations. 
Although the search space differs (token values vs. decoding order), 
the computational burden remains quadratic in $K$, 
making it substantially more expensive than our Order-Token Search, 
which scales only linearly with $K$.

\begin{table*}[t]
\caption{\textbf{4$\times$4 Sudoku result.} We report Sudoku average per-cell accuracy across multiple generation lengths for two base models (LLaDA and LLaDA-1.5). Bolded values indicate the best performance, and underlined values indicate the second best. We treat Sudoku as an outlier relative to the math/coding benchmarks in Table~\ref{tab:performance}. \emph{Reference:} a completely random method would fill in a $4\times4$ Sudoku with $25\%$ of the cells correct in expectation.}
\label{tab:sudoku_only}
\centering
\begin{tabular}{ll|ccccc}
\toprule
 &  & \multicolumn{5}{c}{Sudoku} \\
\cmidrule(lr){3-7}
Model & Method / Seq Len & 64 & 128 & 256 & 512 & Avg \\
\midrule
LLaDA
& Low-confidence        & 8.5 & 11.7 & 6.5 & 5.5 & 8.1 \\
& Low-conf + MV         & 8.9 & 10.3 & 8.5 & 7.2 & 8.7 \\
& Random + MV           & 6.4 & 6.5  & 8.4 & 6.9 & 7.1 \\
& AR                    & 14.7 & 15.9 & 13.6 & 12.0 & \underline{14.1} \\
& AR + MV               & 14.4 & 15.9 & 12.1 & 14.3 & \textbf{14.2} \\
& AR + beam-search      & 13.2 & 17.7 & 10.6 & 8.2  & 12.4 \\
& OTS All Blocks        & 6.5  & 6.9  & 7.5  & 6.9  & 6.9 \\
& OTS Future Blocks     & 7.0  & 7.7  & 7.4  & 6.9  & 7.3 \\
& \textbf{Order-Token Search} & 10.1 & 11.7 & 8.5 & 7.4 & 9.4 \\
\midrule
LLaDA-1.5
& Low-confidence        & 13.5 & 15.6 & 11.4 & 10.3 & 12.7 \\
& Low-conf + MV         & 13.7 & 14.9 & 10.8 & 12.6 & 13.0 \\
& Random + MV           & 11.1 & 11.3 & 9.3  & 9.2  & 10.2 \\
& AR                    & 14.5 & 15.8 & 14.5 & 13.6 & \underline{14.6} \\
& AR + MV               & 14.6 & 16.6 & 13.2 & 13.1 & 14.4 \\
& AR + beam-search      & 17.5 & 16.8 & 13.5 & 11.0 & \textbf{14.7} \\
& OTS All Blocks        & 9.3  & 10.5 & 9.6  & 9.7  & 9.8 \\
& OTS Future Blocks     & 9.6  & 11.1 & 8.3  & 8.5  & 9.4 \\
& \textbf{Order-Token Search} & 13.8 & 16.1 & 11.2 & 9.9 & 12.8 \\
\bottomrule
\end{tabular}
\end{table*}

\subsection{Limitations on Sudoku Benchmarks}
\label{app:discussion-sudoku}

Table~\ref{tab:sudoku_only} indicates that Sudoku behaves qualitatively differently from the other benchmarks.
Across both LLaDA and LLaDA-1.5, performance on Sudoku remains uniformly low across a wide range of decoding strategies, including autoregressive decoding, majority voting, beam search, and our proposed \OURMETHODtext.
Notably, these methods do not exceed a simple random baseline: for a $4\times4$ Sudoku, a uniformly random fill attains $25\%$ expected cell accuracy.

Overall, this pattern suggests that Sudoku performance is primarily limited by how well the underlying model internalizes the task's global, hard constraints, rather than by the choice of inference-time decoding alone.
Sudoku requires enforcing row, column, and subgrid consistency, and such constraint structure may not be reliably captured by token likelihoods.
When the model's scores are weakly correlated with constraint satisfaction, search-based decoding has limited leverage: exploring decoding trajectories mainly reorders and recombines signals already present in the model, without introducing new structural knowledge.

\OURMETHOD is designed to improve reasoning by exploring and pruning \emph{order--token trajectories} under a fixed model.
It is most effective in regimes where the model assigns informative relative likelihoods to intermediate states that reflect partial correctness.
On Sudoku, however, the model appears to assign nearly uninformative or occasionally misleading scores to constraint-violating states, which can cause diverse decoding methods (including AR, beam search, and \OURMETHODtext) to saturate at similarly low accuracy.

We therefore view Sudoku as a boundary case for inference-time structured search: decoding can amplify and select among \emph{existing} reasoning signals, but its gains depend on the presence of reliable constraint-related signal in the model's scoring.
Improving performance on Sudoku likely requires model-level changes, such as explicit constraint modeling or targeted training, rather than more sophisticated decoding alone.

For this reason, we report Sudoku separately and exclude it from aggregate conclusions about reasoning improvements, treating it as a diagnostic case study of model capability on global constraint satisfaction.


\begin{figure}[t]
\centering
\includegraphics[width=0.8\linewidth]{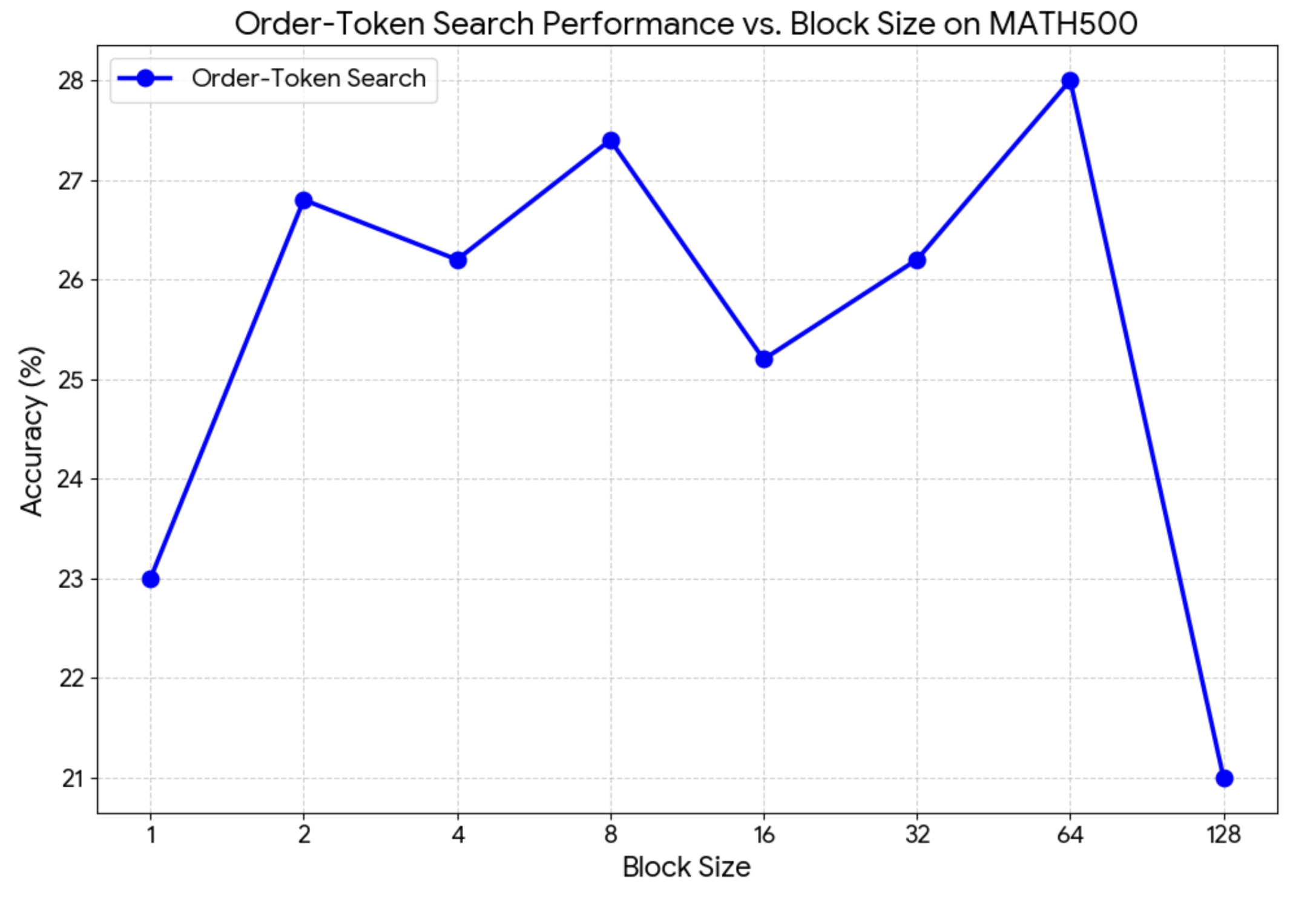}
\caption{\update{\textbf{Effect of block size on OTS accuracy} on MATH500 with generation length 128. Accuracy remains stable for block sizes 2–64, while the degenerate settings of block size 1 and 128 significantly underperform, confirming that block size mainly acts as an efficiency and granularity knob.}}
\label{fig:block-size-ablation}
\end{figure}

\subsection{\update{Sensitivity of Order-Token Search to Block Size}}
\label{app:block-size}

\update{Our scoring function $s(x_t; x_s)$ is explicitly designed to be stable across a range of block sizes. Conceptually, the block size controls a bias--variance trade-off in likelihood estimation. When the block is larger, the model must jointly predict more tokens at once, making each scoring step harder but fewer in number. When the block is smaller, each prediction is easier and closer to the MDM training distribution---where the model typically denoises a limited number of masks at a time---but search is invoked more frequently. In all cases, the score of a candidate is the sum of these incremental block-level log-likelihoods over its full generation path (Eq.~\ref{eq:likelihood-estimate}), so changing the block size simply changes how finely this path-wise likelihood is decomposed, not the underlying distribution being estimated. We therefore view the block size primarily as an efficiency and granularity knob rather than a fragile hyperparameter for the scoring rule itself.}

\update{In practice, we find that \OURMETHOD is not highly sensitive to the exact block size within a reasonable range. On MATH500 with generation length 128, sweeping the block size from $1$ to $128$ yields accuracies between $23.0\%$ and $28.0\%$. Across block sizes $2$--$64$, performance stays in a narrow band around $26.5\%$ (approximately $26.5 \pm 1.5$), and all such settings significantly outperform the degenerate cases of block size $1$ and $128$, where \OURMETHOD either loses the order space entirely (block size $1$) or forces the model to effectively denoise the entire sequence in one shot (block size $128$). This empirical plateau for intermediate block sizes matches the bias--variance trade-off discussed above and supports the view that block size primarily controls the efficiency and granularity of search rather than acting as a delicate tuning parameter. The full sweep is visualized in Figure~\ref{fig:block-size-ablation}.}

\begin{figure*}[t]
    \centering
        \includegraphics[width=1.0\linewidth]{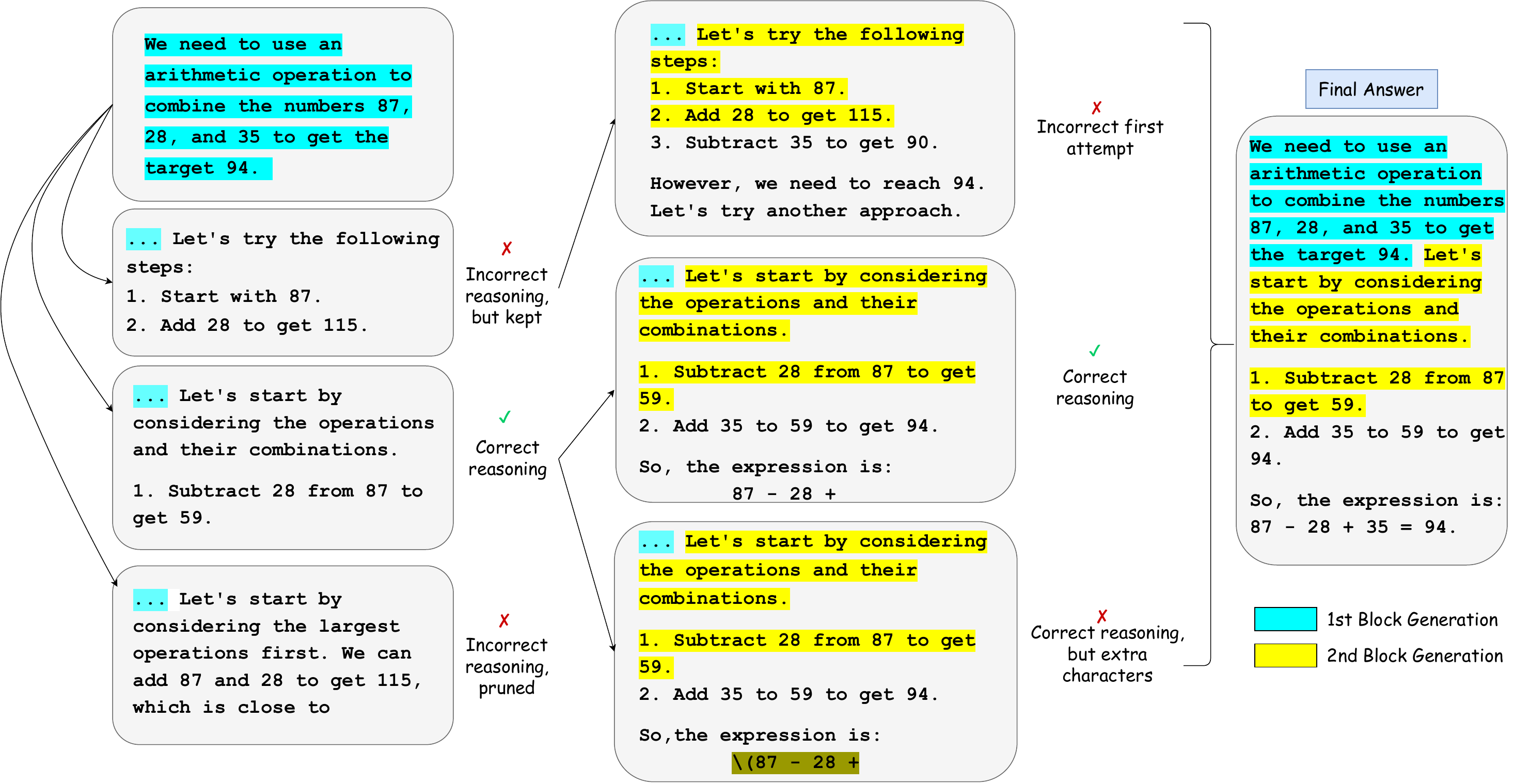}
    \caption{Case study of search trajectories for a sampled Countdown problem. Each box depicts a independently generated candidate sequence with arrows denoting the parent-child relationship in block diffusion. \OURMETHOD evaluates each candidate and decides whether to move forward with its prefix sequence. Its likelihood criterion successfully pruned out inferior candidates that contains incorrect reasoning or syntactical errors, ultimately retaining only high‑scoring candidates that lead to the correct solution.}
    \label{fig:case-study}
\end{figure*}

\subsection{Case Study: A Search Instance} 
\label{app:case-study}

Figure~\ref{fig:case-study} provides a qualitative analysis of \OURMETHODtext's search trajectory on a Countdown task requiring the combination of numbers $(87, 28, 35)$ to reach a target value of $94$. This particular problem exemplifies a case where the low-confidence remasking baseline fails, as it greedily commits to the locally plausible but ultimately incorrect path beginning with $87 + 28 = 115$. However, this path cannot yield the target $94$ using the remaining number $35$, since $115 \pm 35$ results in values ($80$ or $150$) distant from the solution.

\OURMETHOD overcomes this limitation by maintaining multiple candidate paths simultaneously. While the addition-based path is explored, the algorithm also evaluates the alternative $87 - 28 = 59$ trajectory. The dedicated likelihood estimation correctly identifies the subtraction path as superior when $59 + 35$ precisely yields the target $94$. This case demonstrates how \OURMETHODtext's joint exploration of generation orders and token space enables escape from local optima that trap greedy methods, systematically identifying globally correct solutions through parallel hypothesis testing.

\end{document}